\documentclass[runningheads]{llncs}
\usepackage[T1]{fontenc}
\usepackage[table]{xcolor}

\definecolor{codegreen}{rgb}{0,0.5,0}
\definecolor{codeblue}{rgb}{0.25,0.5,0.5}
\definecolor{codegray}{rgb}{0.6,0.6,0.6}
\definecolor{mygray}{gray}{.9}

\usepackage{graphicx}
\usepackage{amsmath}
\usepackage{bm}
\usepackage{pifont}
\usepackage{multirow}
\usepackage{color}
\usepackage[hyphens]{url}
\usepackage{breakurl}

\newcommand{\cmark}{\ding{51}}%
\newcommand{\xmark}{\ding{55}}%
\newlength\savewidth\newcommand\shline{\noalign{\global\savewidth\arrayrulewidth\global\arrayrulewidth 1pt}\hline\noalign{\global\arrayrulewidth\savewidth}}

\begin{document}
\title{Semi-supervised Cell Recognition under Point Supervision\thanks{These authors contributed equally to this work.}}

%
\author{Zhongyi Shui\inst{1,2,*}, Yizhi Zhao\inst{1,2,*}, Sunyi Zheng\inst{2}, Yunlong Zhang\inst{1,2}, Honglin Li\inst{1,2}, Shichuan Zhang\inst{1,2}, Xiaoxuan Yu\inst{1,2},  Chenglu Zhu\inst{2}, Lin Yang\inst{2}}
\authorrunning{Z. Shui et al.}
\institute{College of Computer Science and Technology, Zhejiang University \and
		School of Engineering, Westlake University \\
	\email{yanglin@westlake.edu.cn}}
\maketitle              
\begin{abstract}
Cell recognition is a fundamental task in digital histopathology image analysis. Point-based cell recognition (PCR) methods normally require a vast number of annotations, which is extremely costly, time-consuming and labor-intensive. Semi-supervised learning (SSL) can provide a shortcut to make full use of cell information in gigapixel whole slide images without exhaustive labeling. However, research into semi-supervised point-based cell recognition (SSPCR) remains largely overlooked. Previous SSPCR works are all built on density map-based PCR models, which suffer from unsatisfactory accuracy, slow inference speed and high sensitivity to hyper-parameters. To address these issues, end-to-end PCR models are proposed recently. In this paper, we develop a SSPCR framework suitable for the end-to-end PCR models for the first time. Overall, we use the current models to generate pseudo labels for unlabeled images, which are in turn utilized to supervise the models training. Besides, we introduce a co-teaching strategy to overcome the confirmation bias problem that generally exists in self-training. A distribution alignment technique is also incorporated to produce high-quality, unbiased pseudo labels for unlabeled data. Experimental results on four histopathology datasets concerning different types of staining styles show the effectiveness and versatility of the proposed framework. The code is available at \textcolor{magenta}{\url{https://github.com/windygooo/SSPCR}}.


\end{abstract}
\section{Introduction}
\begin{figure}[t]
	\includegraphics[width=\textwidth]{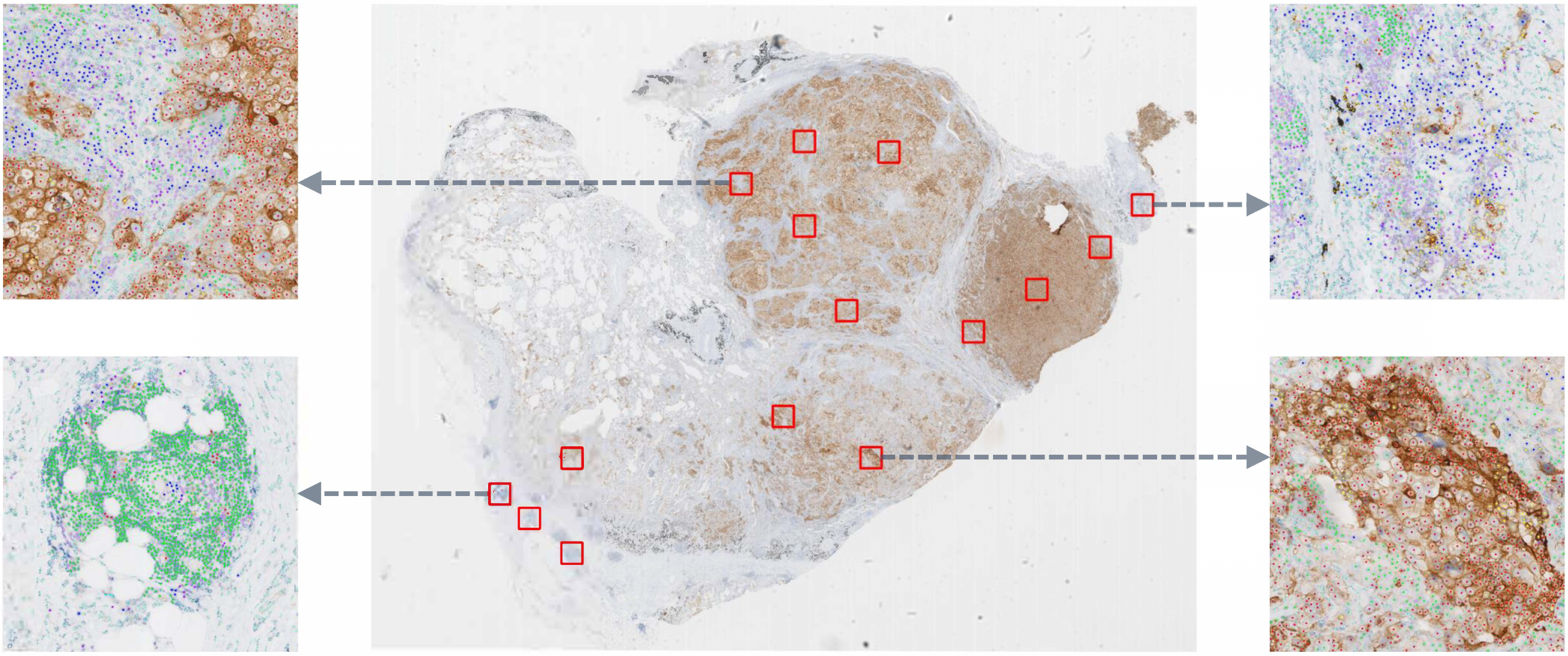}
	\caption{Due to limited human resources in actual scenes, only a small portion of cells, marked by the red boxes, is annotated for developing PCR models. The large number of unlabeled cells outside the boxes contains much valuable information for cell recognition, but is commonly ignored.} \label{fig:background}
\end{figure}

Cell recognition, which aims to localize and classify cells in histopathology images, is fundamental for numerous downstream tasks including whole slide image (WSI) classification \cite{cheng2022artificial}, tumor microenvironment analysis \cite{jiao2021deep} and cancer prognosis prediction \cite{howard2019exploring}. Recently, point-based cell recognition (PCR) has attracted much attention because of its low annotation cost \cite{huang2020bcdata,shui2022end,zhang2022weakly}. In general, the histopathology images for the training of PCR models are cropped from WSIs. As exhaustively labeling hundreds of thousands of cells in a gigapixel WSI is extremely expensive, it is a common practice to crop a few region-of-interest (ROI) patches from WSIs for annotation, leaving a large proportion of cells unused (see Fig.~\ref{fig:background}). Without a shadow of doubt, exploiting these unlabeled cells effectively would improve the performance of full-supervised PCR models. However, under the point annotation setting, the study on how to use these unlabeled cells remains largely under-explored.

Semi-supervised learning (SSL) that intends to use labeled as well as unlabeled data to perform specific learning tasks \cite{van2020survey} provides a shortcut to make full use of cells in WSIs. To the best of our knowledge, there are three papers \cite{bai2021novel,su2023dual,tian2022semi} exploring the pathways of semi-supervised point-based cell recognition (SSPCR) so far. In \cite{bai2021novel}, the model is retrained with the predicted density map for unlabeled images and this progress is repeated for several times. \cite{su2023dual} performs SSL via global consistency regularization and local consistency adversarial learning.  \cite{tian2022semi} utilizes unlabeled samples via location-aware adversarial image reconstruction. However, these three frameworks are all built on density map-based PCR models \cite{ronneberger2015u,xie2018microscopy,zhang2022weakly}, which inevitably suffer from unsatisfactory accuracy, low inference efficiency and extensive, data-specific hyper-parameter tuning due to the need for pre-and post-processing \cite{shui2022end}. To address these problems, recent studies \cite{liang2022end,shui2022end,shui2023deformable,song2021rethinking} propose end-to-end PCR models that can directly output the coordinates and categories of cells, exhibiting superior cell recognition accuracy and efficiency over density map-based PCR models \cite{shui2022end}. However, the existing SSPCR frameworks are incompatible with the end-to-end PCR models.

In this paper, we contribute a SSPCR framework applicable for the state-of-the-art (SOTA) end-to-end PCR models for the first time. Overall, the proposed framework follows the teacher-student self-training paradigm, where the teacher model is updated by the student model in an Exponential Moving Average (EMA) manner and meanwhile provides pseudo point annotations on unlabeled images for the student model training. Moreover, we introduce co-teaching \cite{li2022diverse,liu2022cycle} and distribution alignment \cite{he2021re,chen2022label} techniques to improve the effectiveness of our framework. Extensive experiments on four histopathology datasets with different types of staining styles (HE, Ki-67, PD-L1 and HER2) show that our method improves the performance of the end-to-end PCR models significantly under various labeling ratios.


\begin{figure}[t!]
	\includegraphics[width=\textwidth]{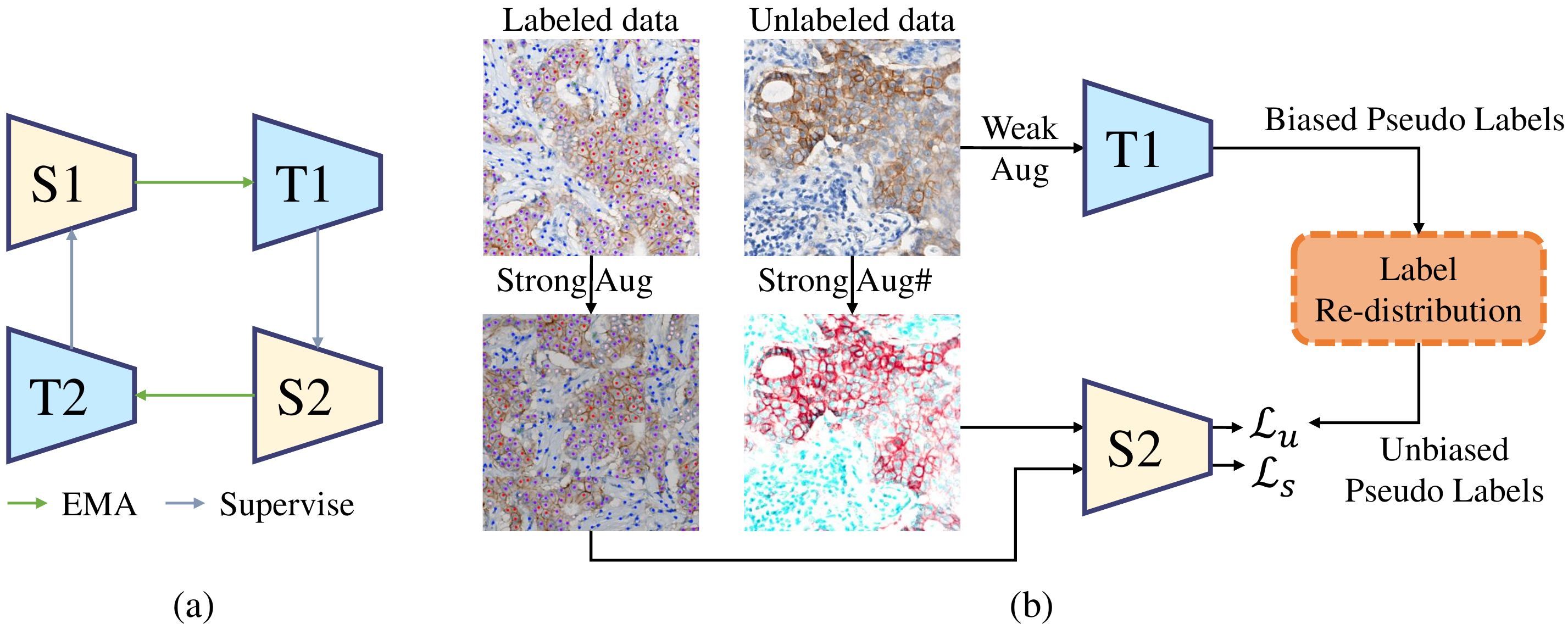}
	\caption{Overview of our proposed SSCPR framework. (a) Schematic of the co-teaching strategy. S and T represent student and teacher models, respectively. (b) Detailed framework. The symbol \# indicates that the strong augmentations applied on labeled and unlabeled images are different. $\mathcal{L}_s$ and $\mathcal{L}_u$ represent the training losses calculated with ground-truth and pseudo labels separately.} \label{fig:framework}
\end{figure}

\section{Semi-Supervised Learning Framework}
In SSPCR, a set of labeled images $\mathcal{D}_l=\{(x_i^l,y_i^l)\}_{i=1}^{N_l}$ and a set of unlabeled images $\mathcal{D}_u=\{x_i^u\}_{i=1}^{N_u}$ are available, where $N_l$ and $N_u$ denote the number of labeled and unlabeled data, respectively. Usually, $N_u \gg N_l$. For each labeled image $x_i^l$, the annotation $y_i^l$ comprises locations and cell categories of all points.

We illustrate the proposed framework in Fig.~\ref{fig:framework}. In the following sections, we first describe the teacher-student mutual learning scheme and then introduce the co-teaching strategy \cite{li2022diverse,liu2022cycle} where two paired teacher-student models are built to provide pseudo labels crossly. Finally, we elaborate how to generate unbiased pseudo labels using the distribution alignment technique \cite{chen2022label,he2021re}.

\paragraph{\textbf{Teacher-student Mutual Learning}}
The proposed framework adopts the pseudo-labeling method to utilize unlabeled samples. As the quality of pseudo labels is critical to the performance of our method, we maintain a teacher model, which can be regarded as a temporal ensemble of the student models at different training iterations, to generate accurate pseudo labels \cite{liu2021unbiased}. To be specific, the student model is optimized by back-propagation with both ground-truth and pseudo labels, whereas the teacher model is slowly updated via Exponential Moving Average (EMA):
\begin{equation}
	\begin{aligned}
		\theta_s &= \theta_s + \eta \frac{\partial \mathcal{L}}{\partial \theta_s} \\
		\theta_t &= \alpha \theta_t + (1-\alpha) \theta_s
	\end{aligned}
\end{equation}
where $\theta_s$ and $\theta_t$ represent the parameters of the student and teacher models, respectively. $\mathcal{L}$ is the training loss and $\eta$ denotes the learning rate. $\alpha \in [0,1]$ is a free parameter controlling the update speed of the teacher model.

In the training process, a mini-batch is composed of $n_l$ labeled and $n_u$ unlabeled images. We apply strong augmentation on the labeled images to increase the diversity of the student model so that the performance of the teacher model can be progressively improved \cite{liu2021unbiased}. The unlabeled images processed by two data augmentations with different strengths are used as input of the student and teacher models, respectively. Concretely, the teacher model takes weakly augmented unlabeled images as input to produce reliable pseudo labels while the strongly perturbed ones are fed into the student model for consistency learning \cite{liu2021unbiased,xie2020unsupervised}.

\paragraph{\textbf{Co-teaching}}
As revealed in \cite{li2022diverse}, the teacher model is prone to make similar predictions with the student model as training proceeds, leading to error accumulation once incorrect pseudo labels are injected. To alleviate this confirmation bias problem, we introduce a co-teaching strategy from \cite{li2022diverse,liu2022cycle}. Specifically, we train two student models (S1, S2) with different initializations simultaneously, each of which maintains its paired teacher model (T1 or T2) via EMA. T1 and T2 are separately used to generate pseudo labels for the training of S2 and S1, which allows the rectification of incorrect pseudo labels when at least one teacher model gives the correct prediction for an unlabeled sample.

\paragraph{\textbf{Unbiased Pseudo Label Generation}}
Class imbalance is a widespread problem in cell recognition applications. It is well known that deep learning models trained on an imbalanced dataset would produce predictions biased toward the dominant categories \cite{he2021re}. Therefore, the class distribution of pseudo labels deviates from the true distribution. Furthermore, the imbalance ratio would further increase if a single confidence threshold is applied upon all categories of predictions to filter out low-quality pseudo labels. Training with such biased pseudo labels can impair the performance of PCR models significantly \cite{liu2021unbiased}.

To generate unbiased pseudo labels using a teacher model, we introduce a distribution alignment technique \cite{chen2022label,he2021re} to customize class-specific thresholds $T=\{t_i\}_{i=1}^C$. Specifically, the thresholds are calculated through the following equations:
\begin{equation}
	\hat{n}_i^u = \frac{N_u}{N_l} \cdot n_i^l,\ i=1,\cdots,C
\end{equation}
where $n_i^l$ is the number of labeled cells of category $i$. $\hat{n}_i^u$ denotes the number of cells predicted as the $i$-th class with confidence scores higher than $t_i$ in the unlabeled data. With the support of $T$, we can align the class distribution of  pseudo labels generated by a teacher model with that of labeled data, which greatly mitigates the side effect of class imbalance. It is worth noting that we refresh $T$ at the start of each epoch to ensure its timeliness.

\paragraph{\textbf{Loss Function}}
The end-to-end PCR methods \cite{liang2022end,shui2023deformable,shui2022end,song2021rethinking} are optimized by minimizing the weighted sum of classification and regression losses under fully supervised conditions. Prior semi-supervised object detection studies \cite{chen2022label,li2022diverse,liu2022cycle,liu2021unbiased} show that locations of pseudo boxes are seriously noisy. We also find this positional noise under the setting of point annotation. Therefore, we only utilize the unlabeled data to supervise the classification learning of the student models. Overall, the loss function of the proposed SSPCR framework is:
\begin{equation}
	\mathcal{L} = \mathcal{L}_s^{cls} + \lambda \mathcal{L}_s^{reg} + \beta \mathcal{L}_u^{cls}
\end{equation}
where $\mathcal{L}_s^{cls}$ and $\mathcal{L}_u^{cls}$ represent the classification loss calculated with labeled and unlabeled data, respectively. $\mathcal{L}_s^{reg}$ denote the regression loss calculated with labeled data. $\lambda$ and $\beta$ are weighting factors. The calculation details about $\mathcal{L}^{reg}$ and $\mathcal{L}^{cls}$ can be found in \cite{shui2023deformable}.

\section{Experiments}

\subsection{Dataset description and experimental settings}
\paragraph{\textbf{Datasets.}}
We conduct experiments on four histopathology datasets with different staining styles (HE, Ki-67, PD-L1 and HER2), where 569466, 138644, 466200 and 833807 cell instances are labeled, respectively. The HE \cite{graham2021lizard,graham2021conic}, Ki-67, PD-L1 and HER2 stained datasets separately contain six, six, ten and six types of cell annotations. More information about these datasets including data sources, cell classes and image resolutions can be found in the supplementary material. We divide each dataset randomly into training, validation and test subsets in a 6:2:2 ratio. The effectiveness of the proposed SSPCR framework is validated with 5\%, 10\%, 15\% and 20\% ground-truth labels available in the training subset.

\paragraph{\textbf{Implementation Details}}
We use DPA-P2PNet \cite{shui2023deformable} with the backbone of ResNet-50 \cite{he2016deep} as our cell recognizer. The AdamW optimizer \cite{loshchilov2017decoupled} with a fixed learning rate of 1e-4 is adopted to optimize the student models. The number of labeled and unlabeled images in a mini-batch is set to 4. Note that only the labeled data are used in the first 50 epochs to pre-train the models. Then, both labeled and unlabeled data are used for training in the rest 150 epochs. We set the EMA rate $\alpha$ to 0.99. In the loss function, $\lambda$ is set to 2e-3, and $\beta$ is set to 1. We apply data augmentation of RandomGridShuffle, RandomHorizontalFlip, RandomVerticalFlip, RandomBrightness and RandomContrast on labeled data to boost the model performance. Two augmentation pipelines with different strengths are constructed for the unlabeled data. The weak one that works for the teacher models only comprises RandomHorizontalFlip, while the strong one customized for the student models is composed of ColorJitter and GaussianBlur \cite{liu2021unbiased}. 

\paragraph{\textbf{Evaluation Metric}}
As in previous PCR works \cite{shui2023deformable,shui2022end,zhang2022weakly}, we use macro-average precision (P), recall (R) and F1 to evaluate all models. A predicted point is regarded as true positive (TP) if it is within the region of a ground truth point with a predefined distance threshold $T_m$. Following \cite{shui2022end,shui2023deformable,zhang2022weakly}, we set $T_m$ to 6 on the HE stained dataset obtained at $20\times$ magnification while 12 on the other three IHC stained datasets collected at $40\times$ magnification. 

\begin{table}[t!]
	\centering	 
	\caption{Experimental results on four histopathology datasets. The labeling ratio indicates the percentage of available ground-truth labels in the training subset. SSL means that the unlabeled training images are exploited with our framework to improve the model performance.}	 
	\small{
		\resizebox{0.9\linewidth}{!}{		
			\begin{tabular}{c|c|c|cccc|cccc}
				\shline
				\multirow{2}{*}{Dataset} & Labeling & 
				\multirow{2}{*}{SSL} & \multicolumn{4}{c|}{Detection} & \multicolumn{4}{c}{Classification} \\				
				\cline{4-11}				 
				& ratio && P & R & F1 & $\Delta$F1 & P & R & F1 & $\Delta$F1 \\
				\shline
				\multirow{8}{*}{HE} & \multirow{2}{*}{$5\%$} & \xmark & $80.00$ & $76.65$ & $78.29$ & \multirow{2}{*}{$\bm{+2.04}$} & $47.34$ & $45.92$ & $46.48$ & \multirow{2}{*}{$\bm{+2.79}$} \\
				~ & ~ & \cmark & $77.02$ & $83.94$ & $80.33$ & ~ & $48.50$ & $50.67$ & $49.27$ & ~ \\
				\cline{2-11}
				~ & \multirow{2}{*}{$10\%$} & \xmark & $79.28$ & $83.77$ & $81.46$ & \multirow{2}{*}{$\bm{+1.35}$} & $51.60$ & $51.35$ & $51.13$ & \multirow{2}{*}{$\bm{+2.35}$} \\
				~ & ~ & \cmark & $80.75$ & $84.97$ & $82.81$ & ~ & $53.85$ & $53.46$ & $53.48$ & ~ \\
				\cline{2-11}
				~ & \multirow{2}{*}{$15\%$} & \xmark & $81.87$ & $83.59$ & $82.72$ & \multirow{2}{*}{$\bm{+1.03}$} & $52.68$ & $53.09$ & $52.84$ & \multirow{2}{*}{$\bm{+2.28}$} \\
				~ & ~ & \cmark & $80.04$ & $87.83$ & $83.75$ & ~ & $54.26$ & $56.30$ & $55.12$ & ~ \\
				\cline{2-11}
				~ & \multirow{2}{*}{$20\%$} & \xmark & $79.55$ & $87.49$ & $83.33$ & \multirow{2}{*}{$\bm{+1.08}$} & $53.73$ & $54.91$ & $54.00$ & \multirow{2}{*}{$\bm{+1.85}$} \\
				~ & ~ & \cmark & $81.59$ & $87.44$ & $84.41$ & ~ & $55.61$ & $56.51$ & $55.85$ & ~ \\
				\cline{2-11}
				~ & $100\%$ && $82.56$ & $89.41$ & $85.85$ && $60.73$ & $60.58$ & $60.38$ & \\
				\shline				 	
				\multirow{8}{*}{Ki-67} & \multirow{2}{*}{$5\%$} & \xmark& $66.57$ & $71.21$ & $68.81$ & \multirow{2}{*}{$\bm{+2.77}$} & $45.13$ & $46.44$ & $45.13$ & \multirow{2}{*}{$\bm{+3.14}$} \\
				~ & ~ & \cmark & $68.57$ & $74.87$ & $71.58$ & ~ & $46.61$ & $50.60$ & $48.27$ & ~ \\
				\cline{2-11}
				~ & \multirow{2}{*}{$10\%$} & \xmark & $70.06$ & $75.72$ & $72.78$ & \multirow{2}{*}{$\bm{+1.28}$} & $47.26$ & $50.99$ & $48.86$ & \multirow{2}{*}{$\bm{+2.10}$} \\
				~ & ~ & \cmark & $71.91$ & $76.36$ & $74.06$ & ~ & $50.46$ & $52.10$ & $50.96$ & ~ \\
				\cline{2-11}
				~ & \multirow{2}{*}{$15\%$} & \xmark & $71.33$ & $75.55$ & $73.38$ & \multirow{2}{*}{$\bm{+1.49}$} & $49.97$ & $50.03$ & $49.48$ & \multirow{2}{*}{$\bm{+2.67}$} \\				
				~ & ~ & \cmark & $71.71$ & $78.32$ & $74.87$ &  & $50.50$ & $54.31$ & $52.15$ & \\
				\cline{2-11}
				~ & \multirow{2}{*}{$20\%$} & \xmark & $70.22$ & $78.25$ & $74.02$ & \multirow{2}{*}{$\bm{+1.75}$} & $49.29$ & $53.24$ & $50.63$ & \multirow{2}{*}{$\bm{+2.18}$} \\
				~ & ~ & \cmark & $72.53$ & $79.30$ & $75.77$ & & $51.05$ & $55.15$ & $52.81$ &  \\
				\cline{2-11}
				~ & $100\%$ && $76.43$ & $81.35$ & $78.81$ && $57.55$ & $60.97$ & $59.08$ & \\
				\shline
				\multirow{8}{*}{PD-L1} & \multirow{2}{*}{$5\%$} & \xmark & $60.24$ & $68.76$ & $64.22$ & \multirow{2}{*}{$\bm{+2.87}$} & $32.11$ & $31.80$ & $29.85$ & \multirow{2}{*}{$\bm{+2.54}$} \\
				~ & ~ & \cmark & $65.47$ & $68.80$ & $67.09$ &  & $34.86$ & $32.50$ & $32.39$ &  \\
				\cline{2-11}
				~ & \multirow{2}{*}{$10\%$} & \xmark & $67.04$ & $70.00$ & $68.49$ & \multirow{2}{*}{$\bm{+2.71}$} & $38.26$ & $35.92$ & $36.03$ & \multirow{2}{*}{$\bm{+3.44}$} \\
				~ & ~ & \cmark & $70.73$ & $71.66$ & $71.20$ &  & $42.22$ & $39.32$ & $39.47$ &  \\
				\cline{2-11}
				~ & \multirow{2}{*}{$15\%$} & \xmark & $61.91$ & $78.67$ & $69.29$ & \multirow{2}{*}{$\bm{+3.15}$} & $36.77$ & $40.48$ & $37.16$ & \multirow{2}{*}{$\bm{+3.33}$} \\
				~ & ~ & \cmark & $72.63$ & $72.24$ & $72.44$ &  & $43.47$ & $41.45$ & $40.49$ &  \\
				\cline{2-11}
				~ & \multirow{2}{*}{$20\%$} & \xmark & $64.70$ & $75.02$ & $69.48$ & \multirow{2}{*}{$\bm{+2.06}$} & $40.02$ & $40.65$ & $39.56$ & \multirow{2}{*}{$\bm{+2.73}$} \\
				~ & ~ & \cmark & $69.31$ & $73.93$ & $71.54$ &  & $41.53$ & $44.10$ & $42.29$ &  \\
				\cline{2-11}
				~ & $100\%$ && $71.16$ & $82.34$ & $76.34$ && $50.70$ & $57.26$ & $53.66$ &  \\
				\shline
				\multirow{8}{*}{HER2} & \multirow{2}{*}{$5\%$} & \xmark& $75.53$ & $76.64$ & $76.08$ & \multirow{2}{*}{$\bm{+1.08}$} & $51.53$ & $52.84$ & $51.95$ & \multirow{2}{*}{$\bm{+1.21}$} \\
				~ & ~ & \cmark & $73.57$ & $81.11$ & $77.16$ &  & $53.82$ & $53.95$ & $53.16$ & \\
				\cline{2-11}
				~ & \multirow{2}{*}{$10\%$} & \xmark & $76.83$ & $80.06$ & $78.41$ & \multirow{2}{*}{$\bm{+1.47}$} & $56.00$ & $55.89$ & $55.52$ & \multirow{2}{*}{$\bm{+1.99}$} \\
				~ & ~ & \cmark & $76.56$ & $83.49$ & $79.88$ &  & $57.19$ & $58.46$ & $57.51$ &  \\
				\cline{2-11}
				~ & \multirow{2}{*}{$15\%$} & \xmark & $77.68$ & $81.80$ & $79.69$ & \multirow{2}{*}{$\bm{+1.48}$} & $57.93$ & $58.04$ & $57.75$ & \multirow{2}{*}{$\bm{+2.46}$} \\
				~ & ~ & \cmark & $79.38$ & $83.03$ & $81.17$ &  & $60.51$ & $60.51$ & $60.21$ &  \\
				\cline{2-11}
				~ & \multirow{2}{*}{$20\%$} & \xmark & $80.02$ & $81.43$ & $80.72$ & \multirow{2}{*}{$\bm{+1.23}$} & $61.92$ & $58.91$ & $60.06$ & \multirow{2}{*}{$\bm{+1.98}$} \\
				~ & ~ & \cmark & $80.97$ & $82.96$ & $81.95$ &  & $64.09$ & $60.87$ & $62.04$ &  \\
				\cline{2-11}
				~ & $100\%$ && $84.51$ & $83.22$ & $83.86$ && $71.15$ & $70.23$ & $70.66$ & \\
				\cline{2-11}
				\shline
	\end{tabular}}}
	\vspace{-1.0em}
	\label{tab:performance}
\end{table}

\subsection{Experimental results}
We present the performance gains brought by the proposed SSL method under different datasets and labeling ratios in Table \ref{tab:performance}. The experimental results show that the F1 score of the cell recognizer is improved by 1-4 points in both detection and classification, which demonstrates the effectiveness and versatility of our method. It is worth noting that the model trained on 10\% and 15\% labeled data using our framework outperforms the full supervised baselines trained on 15\% and 20\% labeled data separately on the HE, Ki-67 and PD-L1 datasets. An interesting finding is that the performance gain varies with the labeling ratio non-monotonically, which could be attributed to the combined effect of quality of pseudo labels and volume of unlabeled data. In general, with the increase of the labeling ratio, the pre-trained models could generate pseudo labels with higher quality.

\begin{table}[t!]
	\centering
	\caption{Performance of our framework with different backbones. The experiments are conducted on the HE stained dataset.}
	\small{
		\resizebox{0.9\linewidth}{!}{		
			\begin{tabular}{c|c|c|cccc|cccc}
				\shline
				\multirow{2}{*}{Backbone} & Labeling & 
				\multirow{2}{*}{SSL} & \multicolumn{4}{c|}{Detection} & \multicolumn{4}{c}{Classification} \\				
				\cline{4-11}				 
				& ratio && P & R & F1 & $\Delta$F1 & P & R & F1 & $\Delta$F1 \\
				\shline
				\multirow{8}{*}{ConvNeXt-B} & \multirow{2}{*}{$5\%$} & \xmark & $78.08$ & $82.68$ & $80.31$ & \multirow{2}{*}{$\bm{+0.85}$} & $47.67$ & $48.28$ & $47.63$ & \multirow{2}{*}{$\bm{+2.18}$} \\
				~ & ~ & \cmark & $77.16$ & $85.60$ & $81.16$ &  & $48.25$ & $51.99$ & $49.81$ &  \\
				\cline{2-11}
				~ & \multirow{2}{*}{$10\%$} & \xmark & $82.26$ & $82.33$ & $82.29$ & \multirow{2}{*}{$\bm{+1.49}$} & $53.95$ & $51.68$ & $52.53$ & \multirow{2}{*}{$\bm{+2.93}$} \\
				~ & ~ & \cmark & $82.24$ & $85.39$ & $83.78$ & & $55.41$ & $55.62$ & $55.46$ & \\
				\cline{2-11}
				~ & \multirow{2}{*}{$15\%$} & \xmark & $79.31$ & $87.68$ & $83.29$ & \multirow{2}{*}{$\bm{+1.41}$} & $53.21$ & $56.60$ & $54.70$ & \multirow{2}{*}{$\bm{+2.39}$} \\
				~ & ~ & \cmark & $82.23$ & $87.32$ & $84.70$ &  & $57.24$ & $57.20$ & $57.09$ & \\				
				\cline{2-11}
				~ & \multirow{2}{*}{$20\%$} & \xmark & $82.78$ & $85.85$ & $84.29$ & \multirow{2}{*}{$\bm{+1.17}$} & $55.55$ & $56.05$ & $55.75$ & \multirow{2}{*}{$\bm{+2.15}$} \\
				~ & ~ & \cmark & $83.20$ & $87.85$ & $85.46$ &  & $58.08$ & $58.05$ & $57.90$ &  \\ 	
				\cline{2-11}	
				~ & $100\%$ & & $86.55$ & $88.35$ & $87.44$ & ~ & $64.51$ & $63.14$ & $63.73$ & ~ \\				
				\shline				 		
				\multirow{8}{*}{ViTDet-B} & \multirow{2}{*}{$5\%$} & \xmark & $76.12$ & $73.70$ & $74.90$ & \multirow{2}{*}{$\bm{+2.42}$} & $38.90$ & $38.18$ & $38.49$ & \multirow{2}{*}{$\bm{+4.82}$} \\
				~ & ~ & \cmark & $73.87$ & $81.11$ & $77.32$ & ~ & $42.30$ & $44.70$ & $43.31$ & ~ \\
				\cline{2-11}				 
				~ & \multirow{2}{*}{$10\%$} & \xmark & $75.98$ & $79.25$ & $77.58$ & \multirow{2}{*}{$\bm{+1.36}$} & $42.66$ & $44.72$ & $43.52$ & \multirow{2}{*}{$\bm{+4.74}$} \\
				~ & ~ & \cmark & $76.82$ & $81.19$ & $78.94$ & ~ & $49.09$ & $47.94$ & $48.26$ & ~ \\
				\cline{2-11}
				~ & \multirow{2}{*}{$15\%$} & \xmark & $77.83$ & $79.55$ & $78.68$ & \multirow{2}{*}{$\bm{+1.55}$} & $46.69$ & $45.17$ & $45.82$ & \multirow{2}{*}{$\bm{+3.62}$} \\
				~ & ~ & \cmark & $77.11$ & $83.60$ & $80.23$ & ~ & $49.69$ & $49.71$ & $49.44$ & ~ \\
				\cline{2-11}
				~ & \multirow{2}{*}{$20\%$} & \xmark & $78.49$ & $80.40$ & $79.44$ & \multirow{2}{*}{$\bm{+1.43}$} & $47.09$ & $46.78$ & $46.75$ & \multirow{2}{*}{$\bm{+3.87}$} \\
				~ & ~ & \cmark & $77.92$ & $84.04$ & $80.87$ & ~ & $50.39$ & $51.18$ & $50.62$ & ~ \\
				\cline{2-11}
				~ & $100\%$ &  & $77.21$ & $88.28$ & $82.38$ & ~ & $51.96$ & $56.80$ & $54.18$ & ~ \\
				\shline
	\end{tabular}}}
	\vspace{-0.9em}
	\label{tab:backbone}
\end{table}

To further validate the versatility of the proposed SSPCR framework, we replace ResNet-50 with another two representative backbones (i.e., ConvNext-B \cite{liu2022convnet} and ViTDet-B \cite{li2022exploring}). Due to the page limitation, we only report the experimental results on the HE stained dataset in Table \ref{tab:backbone}. It can been seen that the proposed SSPCR framework also works well for these two backbones. Surprisingly, we find that though ConvNext-B achieves higher baseline performance than ResNet-50, the increase of classification F1 is even larger using our method with 10\%, 15\% and 20\% labels accessible. This can be explained by better pseudo labels and that the performance may be far from saturation on this dataset. We also notice that ViTDet-B has worse performance compared to the convolutional neural network (CNN) based ResNet-50 and ConvNext-B. This is because vision transformer (ViT) based models have weaker inductive bias than CNNs in modeling visual structures and thus require much more labeled data to learn such bias implicitly \cite{xu2021vitae}. In fact, the requirement for large-scale supervised data limits the application of ViTs in medical image analysis tasks where accurate data annotation takes tremendous effort of experienced doctors. Fortunately, the experimental results show that the proposed SSL method improves the performance of ViTDet-B substantially by 3-5\% on classification F1, which provides a solution to unleash the power of ViTs in the PCR task. The generality of our framework is also verified using another end-to-end PCR model (i.e., P2PNet \cite{song2021rethinking}) as the cell recognizer. In general, the performance of P2PNet is consistently improved using the proposed SSPCR framework. The detailed experimental results can be found in the supplementary material.

\begin{table}[t!]
	\centering
	\caption{Ablation study on teacher-student mutual learning (TSML), co-teaching (CT) and distribution alignment (DA) in the case of 5\% labeled HE data.}
	\small{
		\resizebox{0.8\linewidth}{!}{		
			\begin{tabular}{c|c|c|cccc|cccc}
				\shline
				\multirow{2}{*}{TSML} & \multirow{2}{*}{CT} & 
				\multirow{2}{*}{DA} & \multicolumn{4}{c|}{Detection} & \multicolumn{4}{c}{Classification} \\				
				\cline{4-11}				 
				& && P & R & F1 & $\Delta$F1 & P & R & F1 & $\Delta$F1 \\
				\cline{1-11}
				&&& $80.00$ & $76.65$ & $78.29$ & & $47.34$ & $45.92$ & $46.48$ & \\
				\cmark &&& $75.44$ & $83.17$ & $79.11$ & $\bm{+0.82}$ & $48.78$ & $46.94$ & $46.65$ & $\bm{+0.17}$ \\
				\cmark & \cmark & & $73.59$ & $87.38$ & $79.90$ & $\bm{+1.61}$  & $48.75$ & $49.71$ & $47.62$ & $\bm{+1.14}$ \\
				\cmark & & \cmark & $76.76$ & $83.58$ & $80.02$ & $\bm{+1.73}$ & $48.47$ & $49.17$ & $48.38$ & $\bm{+1.90}$ \\
				\cmark & \cmark & \cmark & $77.02$ & $83.94$ & $80.33$ & $\bm{+2.04}$ & $48.50$ & $50.67$ & $49.27$ & $\bm{+2.79}$ \\
				\shline
	\end{tabular}}}
	\vspace{-1.0em}
	\label{tab:ablation}
\end{table}

\subsection{Ablation Study}
We ablate the effects of teacher-student mutual learning (TSML), distribution alignment (DA) and co-teaching (CT) in the case of 5\%  labeled HE data. As shown in Table.~\ref{tab:ablation}, TSML only promotes the classification F1 by 0.17\%, which can be attributed to the severe class imbalance in pseudo labels. To be specific, the imbalance ratio is 73 in ground-truth labels while 190 in pseudo labels. By further inclusion of CT to mitigate the confirmation bias issue, the model outperforms the baseline by 1.14\% on classification F1. DA leads to a significant improvement (1.90\%) as it generates nearly unbiased pseudo labels, where the imbalance ratio is reduced from 190 to 84. The performance gain reaches 2.79\% by combing these three techniques together.

\section{Conclusion}
In this paper, we design a semi-supervised learning framework adapted to the SOTA end-to-end PCR models for the first time. The proposed SSPCR framework adopts the pseudo-labeling paradigm. Moreover, it incorporates the co-teaching and distribution alignment techniques to overcome the confirmation bias problem and construct unbiased pseudo labels, respectively. Experimental results on four histopathology datasets demonstrate the effectiveness and generality of our proposed framework. The ablation studies validate the efficacy of the framework components. In our future work, we will explore the solutions to promote the localization accuracy of the cell recognizer via unlabeled data.

\bibliographystyle{splncs04}
\bibliography{mybib}

\end{document}